%% file: prob-trees.tex
\documentclass[accepted]{article}

\usepackage{microtype}
\usepackage{graphicx}
\usepackage{subcaption}
\usepackage{booktabs} 
\usepackage{rotating}
\usepackage{adjustbox}
\usepackage{tabularx}
\usepackage{dsfont}
\usepackage{multirow}
\usepackage{multicol}
\usepackage{natbib}

\usepackage{hyperref}
\usepackage{xspace}

\usepackage{icml2023}

\usepackage{amsmath}
\usepackage{amssymb}
\usepackage{mathtools}
\usepackage{amsthm}

\usepackage[capitalize,noabbrev]{cleveref}

\theoremstyle{plain}

\theoremstyle{definition}

\theoremstyle{remark}

\icmltitlerunning{Joint Probability Trees}

\input{makros}

\begin{document}

\graphicspath{{img/}}

\twocolumn[
\icmltitle{Joint Probability Trees}



\icmlsetsymbol{equal}{*}

\begin{icmlauthorlist}
\icmlauthor{Daniel Nyga}{iai,equal}
\icmlauthor{Mareike Picklum}{iai,equal}
\icmlauthor{Tom Schierenbeck}{iai}
\icmlauthor{Michael Beetz}{iai}
\end{icmlauthorlist}

\icmlaffiliation{iai}{Institute for Artificial Intelligence, University of Bremen, Germany}

\icmlcorrespondingauthor{Daniel Nyga}{nyga@cs.uni-bremen.de}

\icmlkeywords{Machine Learning, ICML}

\vskip 0.3in
]



\printAffiliationsAndNotice{\icmlEqualContribution} 


\newcommand{\jpt}{\textsc{JPT}\xspace}
\newcommand{\jpts}{\textsc{JPT}s\xspace}
\newcommand{\pgm}{\textsc{PGM}\xspace}
\newcommand{\pgms}{\textsc{PGM}s\xspace}



\renewcommand{\sectionautorefname}{Section}
\renewcommand{\subsectionautorefname}{Section}
\renewcommand{\subsubsectionautorefname}{Section}
\newcommand{\algorithmautorefname}{Algorithm}


\begin{abstract}
    \input{abstract.tex}
\end{abstract}

\renewcommand{\cite}{\citet}

\section{Introduction}

\input{intro.tex}
\label{sec:intro}

\section{Joint Probability Trees}
\label{sec:jpts}

\input{jpts.tex}

\section{Experiments}
\label{sec:exp}

\input{experiments.tex}
\section{Discussion}
\label{sec:discussion}


\input{desiderata.tex}

\section{Related Work}
\label{sec:related}

\input{related-work.tex}

\section{Conclusions}
\label{sec:conclusions}

\input{conclusions.tex}

\bibliography{literature}
\bibliographystyle{icml2023}

\onecolumn
\section{Appendix}
\input{appendix.tex}







\end{document}

%% file: abstract.tex
We introduce Joint Probability Trees (JPT), a novel approach that makes 
learning of and reasoning about joint probability distributions 
tractable for practical applications. JPTs support both symbolic and 
subsymbolic variables in a single hybrid model, and they do not rely on 
prior knowledge about variable dependencies or families of 
distributions. JPT representations build on tree structures that 
partition the problem space into relevant subregions that are elicited 
from the training data instead of postulating a rigid dependency model 
prior to learning. Learning and reasoning scale linearly in JPTs, and 
the tree structure allows white-box reasoning about any posterior 
probability $P(Q|E)$, such that interpretable explanations can be 
provided for any inference result. Our experiments showcase the 
practical applicability of JPTs in high-dimensional heterogeneous 
probability spaces with millions of training samples, making it a 
promising alternative to classic probabilistic graphical models.


%% file: intro.tex
\begin{figure}
    \centering
     \begin{subfigure}[b]{.49\columnwidth}
         \centering
         \includegraphics[width=\textwidth]{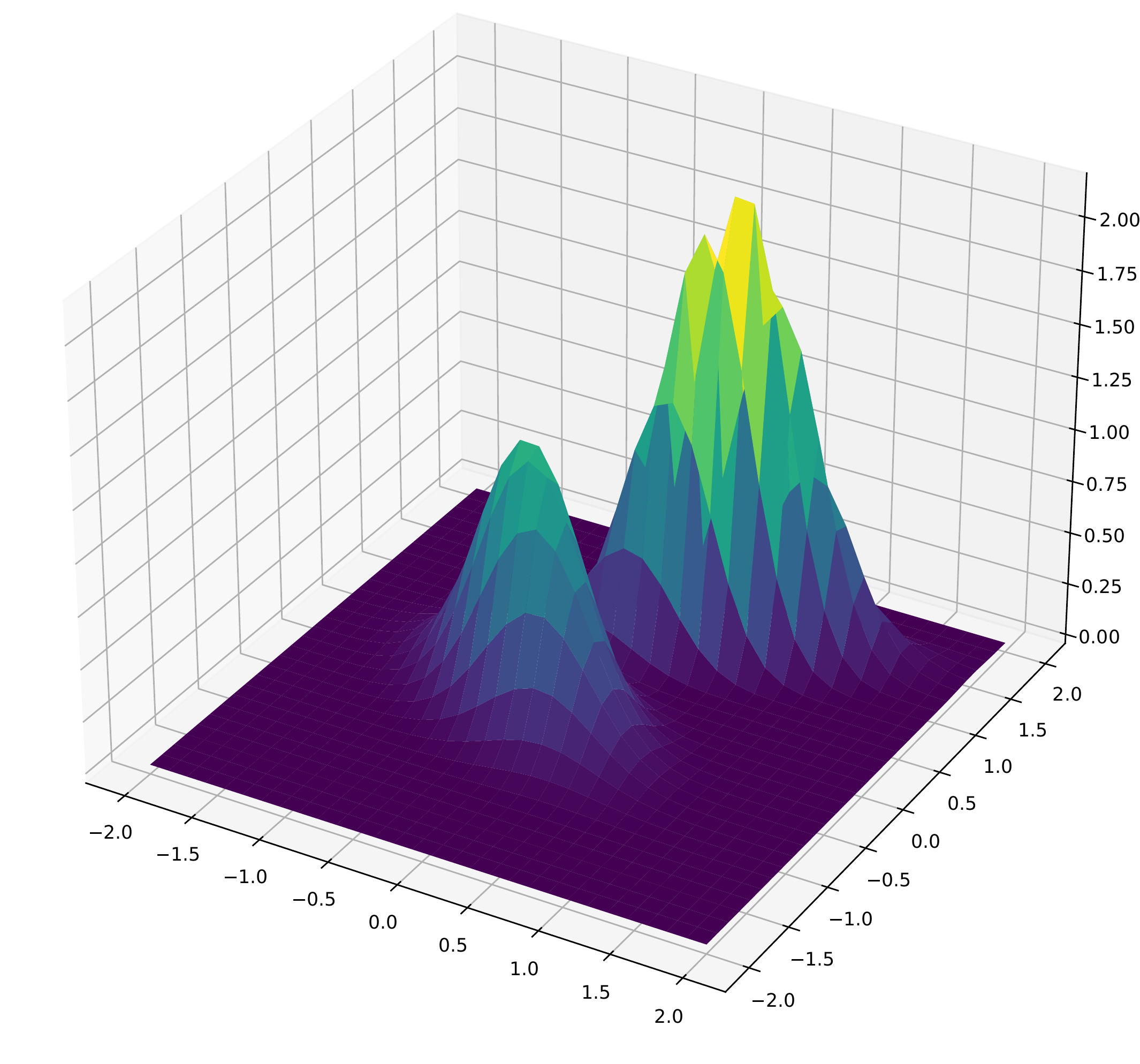}
         \subcaption{The ground truth distribution}
         \label{fig:gaussian-examplea}
     \end{subfigure}
     \begin{subfigure}[b]{.49\columnwidth}
         \centering
         \includegraphics[width=\textwidth]{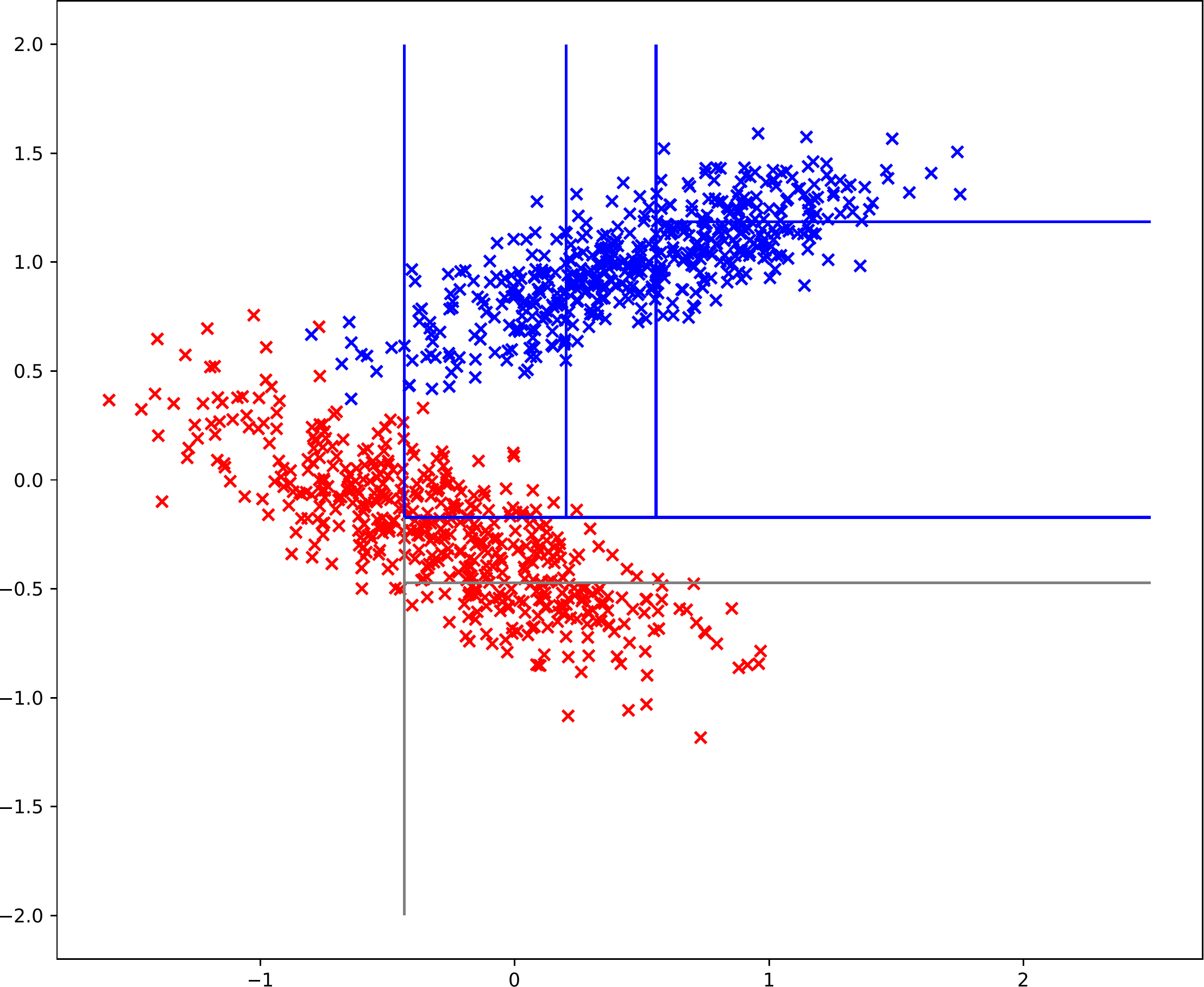}
         \subcaption{Scatterplot of the sample data}
         \label{fig:gaussian-exampleb}
     \end{subfigure}
    \begin{subfigure}[b]{.49\columnwidth}
         \centering
         \includegraphics[width=\textwidth]{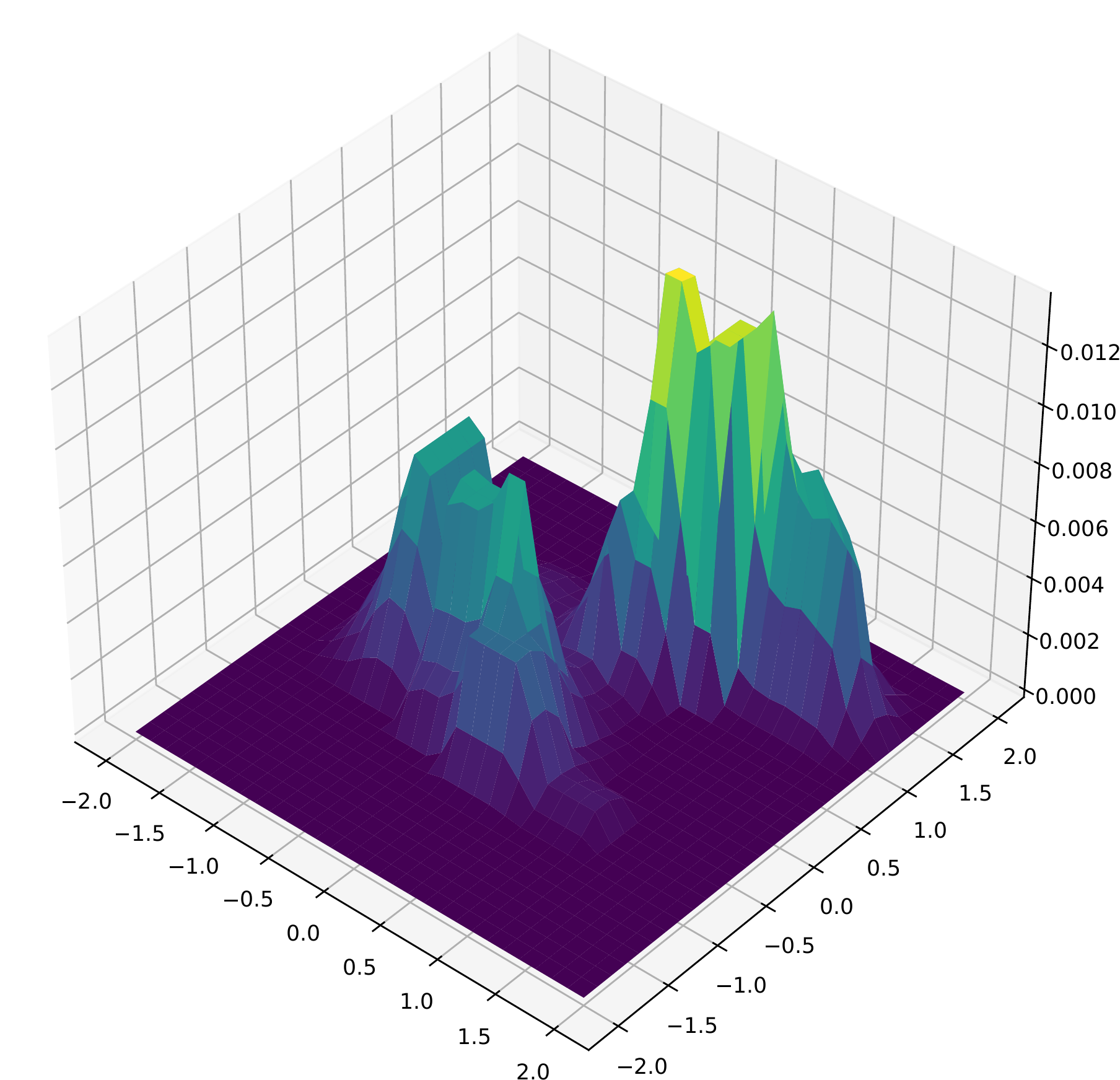}
         \subcaption{The marginal $P(X,Y)$\\ represented by the JPT in (d)}
         \label{fig:gaussian-examplec}
     \end{subfigure}
     \begin{subfigure}[b]{.49\columnwidth}
         \centering
         \includegraphics[width=\textwidth]{gaussian-jpt}
         \subcaption{The tree structure of the corresponding JPT}
         \label{fig:gaussian-exampled}
     \end{subfigure}

    \caption{Example of a joint probability distribution of two numeric 
    variables $(X, Y)$ and one symbolic variable (color). Enlarged
    versions of the figures can be found in the supplementary material.}

    \label{fig:gaussian-example}
\end{figure}

Joint probability distributions offer a wide range of high-potential 
applications in engineering, science, and 
technology~\citep{chater06cognition,griffiths08bayesian,knill04bayesianbrain}. 
Besides families of continuous distributions, probabilistic graphical 
models~(\pgms), such as Bayesian networks and Markov random 
fields~\citep{koller2009probabilistic}, are the de-facto standard in 
probabilistic knowledge representation. They provide graph-based 
languages to model dependencies and independencies of variables, and 
local joint or conditional distributions that quantify the statistical 
dependencies. However, the practical applicability of PGMs suffers from 
the representational and computational complexity of learning and 
reasoning. Exponential runtime for learning and reasoning often can 
only be avoided by introducing strong independence assumptions that 
must be known prior to learning and may turn out to be too great 
simplifications of a model to be of practical 
use~\citep{besag1975statistical,jain2012phd}. As a simple example, 
consider a probability space $\langle X,Y,C\rangle$ of two numeric 
variables, $X$ and $Y$, and one symbolic variable $C$, 
$\textit{dom}(C)=\{\textit{Red},\textit{Blue}\}$ as illustrated in 
Figures~\ref{fig:gaussian-examplea}~and \ref{fig:gaussian-exampleb}. 
Let the symbolic values \textit{Red} and \textit{Blue} demaracate two 
clusters that are approximately normally distributed. Classic methods 
for density estimation postulate a mathematical model and apply the 
maximum likelihood and expectation/maximization principles  in order to 
find the model parameters that fit the data best. However, this 
learning process is expensive since the unconstrained parameter space 
is huge and most learning methods do not exploit the structure of the 
training data and their underlying distribution.

In this paper, we introduce \emph{Joint Probability Trees}~(JPT), a 
novel approach to represent, learn, and reason about uncertain 
knowlegde acquired from data. JPTs are model-free shallow deterministic 
probabilistic circuits and they exploit the structure of the data to be 
learnt from in a greedy fashion in order to construct a tree structure 
that partitions the problem space recursively into subspaces. In its 
leaves, marginal distributions, represented by cumulative distribution 
functions (CDFs) over all variables in the respective subspace are 
maintained, which can be superimposed in order to obtain a sound and 
globally consistent posterior belief. As opposed to most PGMs, 
dependencies among variables in JPTs do not need to be known at design 
time and only very mild assumptions about models are required. 
Figure~\ref{fig:gaussian-exampled} shows such a tree structure that has 
been learnt from the data in Figure~\ref{fig:gaussian-exampleb}, and 
the marginal distribution in Figure~\ref{fig:gaussian-examplec} shows 
indeed close resemblance to the ground truth distribution in 
Figure~\ref{fig:gaussian-examplea}. JPTs allow the computation of any 
posterior distribution in a transparent and explainable way, such that 
the rationale for any inference result can be provided in a 
human-interpretable form.

The contributions of this paper are the following. First, we formally 
introduce the concept of Joint Probability Trees as a novel framework 
for knowledge representation and reasoning in probabilistic hybrid 
domains and we present algorithms for the learning of and reasoning 
about JPTs. Second, we adapt the concept of quantile-parameterized 
distributions for the efficient learning of univariate continuous 
distributions without prior assumptions about their functional forms, 
and third, we investigate and showcase the performance of JPTs 
empirically on publicly available datasets. 


%% file: jpts.tex
In this section, we introduce the concept of Joint Probability Trees 
(JPTs) more formally. Let us denote the $D$-dimensional problem space 
under consideration as $X=\langle X_1,\ldots,X_D\rangle$, where $X$ is 
a vector of random variables $X_i$, whose domains we denote by 
$\textit{dom}(X_i)$. The set of possible worlds, i.e. all possible 
complete variable assignments, is denoted by $\Xcal$, and a specific 
assignment to all variables in $X$, by $x=\langle 
x_1,\ldots,x_D\rangle$, where $x\in\Xcal$. As a representational 
formalism, JPTs make use of tree-like structures like classification 
and regression trees~\citep{breiman2017classification}. Trees have a 
couple of desirable properties that put themselves forward to be used 
as a knowledge representation formalism. Most notably, they (1) are 
simple to understand and interpret, (2) can be thought of as white box 
models that foster explainable decision making, (3) are compact and 
sparse representations that allow efficient learning and reasoning. 
These features are leveraged through the model structures implementing 
a recursive partition of the problem space under consideration. Let $T: 
\Xcal\mapsto\Lambda$ be a tree-like structure that associates an input 
sample $x\in\Xcal$ with one of its leaves 
$\Lambda=\{\lambda_1,\ldots,\lambda_N\}$. 
The recursive partitioning through a tree structure guarantees that $T$
is exhaustive and mutually exclusive.



    


As a consequence, the result of the application of $T$, $T(x)$, can be 
treated as a random variable that indicates which leaf $\lambda$ an 
arbitrary sample $x$ will be associated with. We thus introduce an 
auxiliary variable $L$, $\textit{dom}(L)=\Lambda$, that extends the 
problem space $X$ by $L$ and hence forms a new probability space 
$X'=\left\langle X,L\right\rangle$. Applying the law of total 
probability, we can marginalize the prior probabilities of $X$ over 
$X'$ by 

\begin{align}
    P(X=x)=\sum_{\lambda\in\Lambda}\Pcond{X}{L=\lambda}\P{L=\lambda}.
\end{align} The leaf priors $\P{L=\lambda}$ can be easily obtained 
from the portions of training data that are covered by the respective 
leaf during training and can be permanently stored in the leaf data 
structure. The leaf-conditional distributions 
$\Pcond{X_1,\ldots,X_D}{L=\lambda}$, however, are more difficult to 
represent as they still comprise the joint distributions over $X$. 
Here, we introduce the na\"ive Bayes assumption that postulates 
conditional independence among all $X_i$, once $L$ is known. This 
assumption is reasonable as the leaves in $T$ represent contiguous 
subregions in $\Xcal$ that are formed during learning by minimizing the 
mutual information of variables (or maximizing information gain, 
respectively), which can be proven equivalent to statistical 
dependence~\citep{murphy22pml}. Assuming independence of variables in 
the leaf nodes in turn allows to represent the priors over $X_i$ 
conditioned by the respective subspace in every leaf node of $T$ in an 
extremely compact way: 

\begin{align}
    P(X=x)=\sum_{\lambda\in\Lambda}\P{L=\lambda}\prod_i\Pcond{X_i=x_i}{L=\lambda}.\label{eq:jpt}
\end{align}
We call (\ref{eq:jpt}) the JPT distribution, which can be used in order 
to compute the marginal probability of a possible world $x\in\Xcal$. 
It can be regarded as a mixture model, whose mixing coefficients are
represented by the leaf priors, and the local distributions are given
by the priors in the respective leaves.


\subsection{Reasoning in Joint Probability Trees}


Computing the marginalization over the tree leaves in (\ref{eq:jpt}) 
may seem like unnecessary effort as, once $x$ is known, $\lambda$ is 
fully determined by $T$. However, mixing the leaf distributions becomes 
inevitable if the values of only a fraction of variables $E\subset X$ 
are known in advance and the posterior probability of a subset 
$Q\subseteq X$, $\Pcond{Q}{E}$, needs to be computed. 
Extending~(\ref{eq:jpt}) to canonical posterior inference can be 
achieved in a straightforward way by introducing background evidence to 
the JPT distribution:

\begin{align}
    \Pcond{q}{e}=\sum_{\lambda\in\Lambda}\Pcond{\lambda}{e}\prod_i\Pcond{q_i}{\lambda,e_i},\label{eq:jpt-posterior}
\end{align}
where $q=\langle q_1,\ldots,q_D\rangle$ is a vector of values of the 
query variables and $e=\langle e_1,\ldots,e_D\rangle$ is a vector of 
constraints $e_i$ for the variables $X_i$. Note that, because of the 
na\"ive Bayes assumption, $q_i$ only depends on the constraints on the 
respective same variable $X_i$, i.e. $e_i$, instead of the entire 
vector $e$. 
The most interesting part of the posterior in (\ref{eq:jpt-posterior}) 
is the factor $\Pcond{\lambda}{e}$, the probability that a sample 
satisfying the constraints $e$ will be associated with leaf $\lambda$. 
While the pure leaf priors $\P{\lambda}$ can be obtained from their 
associated data portions, the conditional $\Pcond{\lambda}{e}$ is 
slightly more complex to compute. However, the tree structure of $T$ 
can be exploited as an efficient approximation as follows. According to 
Bayes' theorem, 
$\Pcond{\lambda}{e}\propto\Pcond{e}{\lambda}\cdot\P{\lambda}$ holds. We 
can thus compute the distribution $\Pcond{L}{e}$ by evaluating 
$\Pcond{e}{\lambda}\cdot\P{\lambda}$ for every $\lambda$ and 
normalizing the results to form a proper distribution. 
$\Pcond{e}{\lambda}$ can in turn be factorized according to 
$\prod_i\Pcond{e_i}{\lambda}$ due to the conditional independence 
assumption, which corresponds to the prior distributions stored in 
every leaf of $T$. A significant performance gain can be achieved by 
testing both $q$ and $e$ against the path conditions of every leaf 
$\lambda$. If either of the two violates a path condition, the entire 
subtree can be pruned for a specific reasoning problem.



\subsection{Learning of Joint Probability Trees}


In order to learn a JPT from data, we propose a variant of the popular 
and well-known tree learning methods, which we introduce in this 
section. C4.5~\citep{quinlan1993c4} and CART~\citep{breiman1984cart} 
are very successful variants of induction algorithms 
for classification and regression trees. Essentially, a tree is being built by splitting 
the input data into subsets constituting the input data for the 
successor children. The splitting consists of a variable or a 
variable/value pair that marks a pivot position in the observable 
feature space optimizing some criterion of impurity with respect to the 
split data sets. The split constitutes the root node of the tree and 
the process is repeated on each of the child subsets recursively and 
terminates when the impurity with respect to the target variables 
within a subset at a node is minimal or when splitting no longer 
reduces the impurity. The result of the procedure is a tree structure 
$T$, whose inner nodes represent \textit{decision nodes} at which a 
single input vector $x$ is evaluated according to the pivot variable 
attached to the respective node and the subsequent child node to 
proceed with is determined by the value of that variable. If the child 
node does not have any further children, a \textit{terminal node} is 
reached in which the predictive values of the target variables are 
stored, which constitutes the return value $T(x)$ of the tree. For
a more detailed discussion of tree learning we refer to~\cite{loh2014fifty}.
 
  


\paragraph{Generative learning} Ordinary classification and regression 
tree learning is defined over dedicated feature (input) and target 
(output) variables of the problem space and therefore it is also called 
a discriminative learning setting. In each node, the best possible 
split of the remaining data is looked for in the set of feature 
variables and their domains, and every possible split is evaluated with 
respect to its potential to reduce the impurity of the target 
variables. Although discriminative learning of JPTs is also possible, 
we focus here on the generative learning case. In contrast to 
discriminative CART learning, in JPTs, the entire set of variables $X$ 
functions both as feature and as target variables. This means that in 
every decision node during the learning process, the node impurity is 
evaluated with respect to all available variables $X_i$, and all 
variables $X_i$ are considered as potential split candidates. 

    


As JPTs support both symbolic and numeric variables, a measure of 
impurity is needed to account for this hybrid character of the model. 
Here, the difficulty arises that typical error measures for numerical 
data, e.g. the mean squared error (MSE) and impurity measures for 
symbolic data, e.g. entropy, reside in different and incompatible value 
ranges. In order to harmonize the two worlds of symbolic and 
subsymbolic impurity, we propose a combined measure of normalized, 
relative impurity improvement as follows.
For a distribution over a symbolic random variable $X$, its 
\textit{entropy} is defined by 
$H(P(X))=-\sum_{x\in\textit{dom}(X)}P(x)\log P(x)$. In CART learning, a 
possible split is considered better the more it reduces the expected 
entropy over the children induced by the split. As a multinomial 
variable has its highest entropy in the uniform distribution, we can 
normalize the entropy with respect to the maximal entropy a 
distribution of the same domain size can have, i.e., 
$H(\mathcal{U}(X))$, where $\mathcal{U}(X)$ denotes the uniform 
distribution over $X$. We can thus define the relative entropy of a 
distribution over $X$ as 
$H_\text{rel}(P(X))=H(P(X))/H(\mathcal{U}(X))$. Likewise, the impurity 
of a numeric variable, typically measured by the MSE within a 
population, can be normalized through the percentage by which the MSE 
would be reduced by a split, which we denote by 
$\textit{MSE}_\text{rel}(X)$. The two measures of impurity improvement 
of symbolic and subsymbolic variables are combined by a weighted average
to form the total impurity improvement $I$ over a data set $\Dcal$ when
the split of the data is performed on the variable $X_i$,

\newcommand{\symvars}{X_{\text{sym}}}
\newcommand{\numvars}{X_{\text{num}}}
\begin{align}
    I(\Dcal,X_i)=&\frac{1}{|\symvars|^2}\sum_{X_j\in\symvars}H_\text{rel}(P_{\Dcal,X_i}(X_j))\\
    &+\frac{1}{|\numvars|^2}\sum_{X_j\in\numvars}\textit{MSE}_\text{rel}(P_{\Dcal,X_i}(X_j)),
\end{align}

where $\numvars$ and $\symvars$ denote the sets of numeric and symbolic 
variables in $X$, respectively, and $P_{\Dcal,X_i}(X_j)$ denotes the 
distribution over $X_j$ induced by the data set $\Dcal$ when split at 
variable $X_i$.

When the splitting criterion does not lead to an improvement of the 
impurity within a node or some learning threshold is reached, a 
terminal node is generated by the learning algorithm. Terminal nodes in 
JPTs hold the marginal univariate distributions over all variables in 
$X$ that can be induced by the data in the current node. In the 
next section, we will discuss in greater detail how these 
distributions can be represented and learnt efficiently.


\paragraph{Discriminative learning} JPTs can also be learnt in a 
discriminative fashion. Discriminative learning can be advantageous, 
when it is possible to commit to dedicated sets of input and output 
variables. In such cases, the learning process can be more efficient 
and the learnt model can be more compact and more accurate than its 
generative counterpart. The JPT learning algorithm reduces to ordinary 
CART learning, when the set of variables is split into dedicated 
feature and target variables.

\paragraph{Structure learning} In ordinary PGMs like Bayesian or Markov 
networks, learning the structure of the graphs, i.e. the dependency 
model of the variables under consideration, represents a difficult 
learning problem on its own. As most of the learning methods in PGMs 
make strong assumptions about a fixed graphical model, learning the 
network structure is typically a problem even harder than learning the 
model parameters alone~\citep{koller2009probabilistic}. It is important 
to note that, in JPTs, the model structure $T$ is elicited from the 
data distribution during the learning process and thus $T$ encodes the 
dependencies among the variables. This entails two essential benefits: 
First, no additional computational effort needs to be carried out in 
order to obtain the graphical model, and second, no prior knowledge 
about the domain of discourse must be incorporated prior to learning. 
In particular, in data mining and knowledge discovery applications, this 
is a highly desirable property of a learning algorithm.


\subsection{Example}
\label{sec:example}


Let us illustrate the concept of JPTs by means of the example we 
already presented in the Introduction, see 
Figure~\ref{fig:gaussian-example}. The JPT that has been acquired from 
these data is shown in Figure~\ref{fig:gaussian-exampled}. Every leaf in 
the tree structure corresponds to one rectangular subregion in the 
partition of the problem space in Figure~\ref{fig:gaussian-exampleb}. 
Although the JPT learning does not make any assumptions about the 
functional form of the distribution, the two clusters in the 
distribution are represented reasonably well. Every leaf has also 
attached the prior distributions over all three variables in the 
respective subregion. Visualizations of the distributions are shown as 
CDF plots for the distributions over the two numeric variables and as 
histograms for the distributions over the symbolic variable. For better 
readability, we put an enlarged version of each of the images into 
the supplementary material accompanying this paper.

\section{Learning \& Reasoning in Continuous Domains}
\label{sec:subsymreason}

\newcommand{\Dtilde}{\widetilde{\Dcal}}

\newcommand{\fringe}{\textit{fringe}\xspace}
\newcommand{\node}{\textit{node}\xspace}
\newcommand{\closed}{\textit{closed}\xspace}
\newcommand{\steps}{\textit{steps}\xspace}
\newcommand{\module}{\textit{mod}\xspace}
\newcommand{\modules}{\textit{modules}\xspace}
\newcommand{\newnodes}{\textit{newnodes}\xspace}

\newcommand{\forward}{\textsc{CDF-Learn}\xspace}
\newcommand{\backward}{\textsc{Backward}\xspace}
\newcommand{\makenode}{\textsc{Make-Node}}
\newcommand{\hashmap}{\textsc{Hash-Map}}
\newcommand{\isempty}{\textsc{Empty?}}
\newcommand{\insertnode}{\textsc{Append}}
\newcommand{\insertall}{\textsc{Append-All}}
\newcommand{\genplan}{\textsc{Generate-Plan}}
\newcommand{\parse}{\textsc{NL-Parse}}
\newcommand{\pop}{\textsc{Pop}}
\newcommand{\nextmod}{\textsc{Next-Module}}
\newcommand{\expand}{\textsc{Expand}}
\newcommand{\state}{\textsc{State}}
\newcommand{\Copy}{\textsc{Copy}}
\newcommand{\Next}{\textsc{Next}}


Committing oneself to a specific functional form of the probability 
density function (PDF) beforehand and hence to its respective 
model-tailored learning procedures (e.g. a normal distribution with its 
mean and covariance) as required by many state-of-the-art methods, may 
be subject to misrepresent the underlying data. More complex models to 
represent more sophisticated distributions, as provided by Gaussian 
Mixture Models or kernel-based methods tackle this problem but typically 
require iterative methods like expectation maximization (EM) as they 
cannot be optimized in a closed form.


\subsection{Quantile-parameterized Distributions}
\label{sec:qpd}


\cite{keenlin2011quantile} introduce the concept of 
\textit{quantile-parameterized distributions} (QPD), which we adapt in 
this paper for the purpose of representing and acquiring continuous 
univariate probability distributions. In QPDs, the \textit{cumulative 
distribution function} (CDF) is represented and learned instead of the 
PDF. The CDF, in turn, is the integral over the PDF and represents the 
$\gamma$-quantile probabilities, $\P{X\leq x_\gamma}$, of the 
distribution. The principal advantage of learning the CDF over learning 
the PDF is that \textit{any} CDF can be easily learnt from data as 
follows. Let $\Dcal=\left\{d_1,\ldots,d_N\mid i<j\Rightarrow d_i\leq 
d_j\right\}$ denote the sorted, indexed set of 1-dimensional data 
samples from the continuous domain under consideration. We can then 
construct a dataset $\Dtilde=\left\{\langle 
d_i,\gamma_i\rangle\right\}$ where 
$\gamma_i=\frac{1}{\left|\Dcal\right|}\left|\left\{d\mid d\in\Dcal, 
d\leq d_i\right\}\right|=\frac{i}{\left|\Dcal\right|}$ is the 
$\gamma_i$-quantile of the data set $\Dcal$. $\Dtilde$ serves as 
training data for a supervised regression task, the result of which 
corresponds to the CDF $F(x)$ of the desired probability distribution. 
In principle, any regression model can be used to fit and 
represent the CDF. In order to approximate the CDF of a distribution, 
we propose the use of piecewise linear functions (PLF). A PLF $f(x)$ is 
a function defined on a finite number of intervals in \R, each of which 
has a linear function $f_i$ attached. The set of intervals partition 
the domain of the function. The function value of $f$ at a particular 
point $\xhat$ is given by the value of the function $f_i$ whose 
attached interval encompasses $\xhat$. 

\begin{figure}[tb]
    \includegraphics[width=\columnwidth]{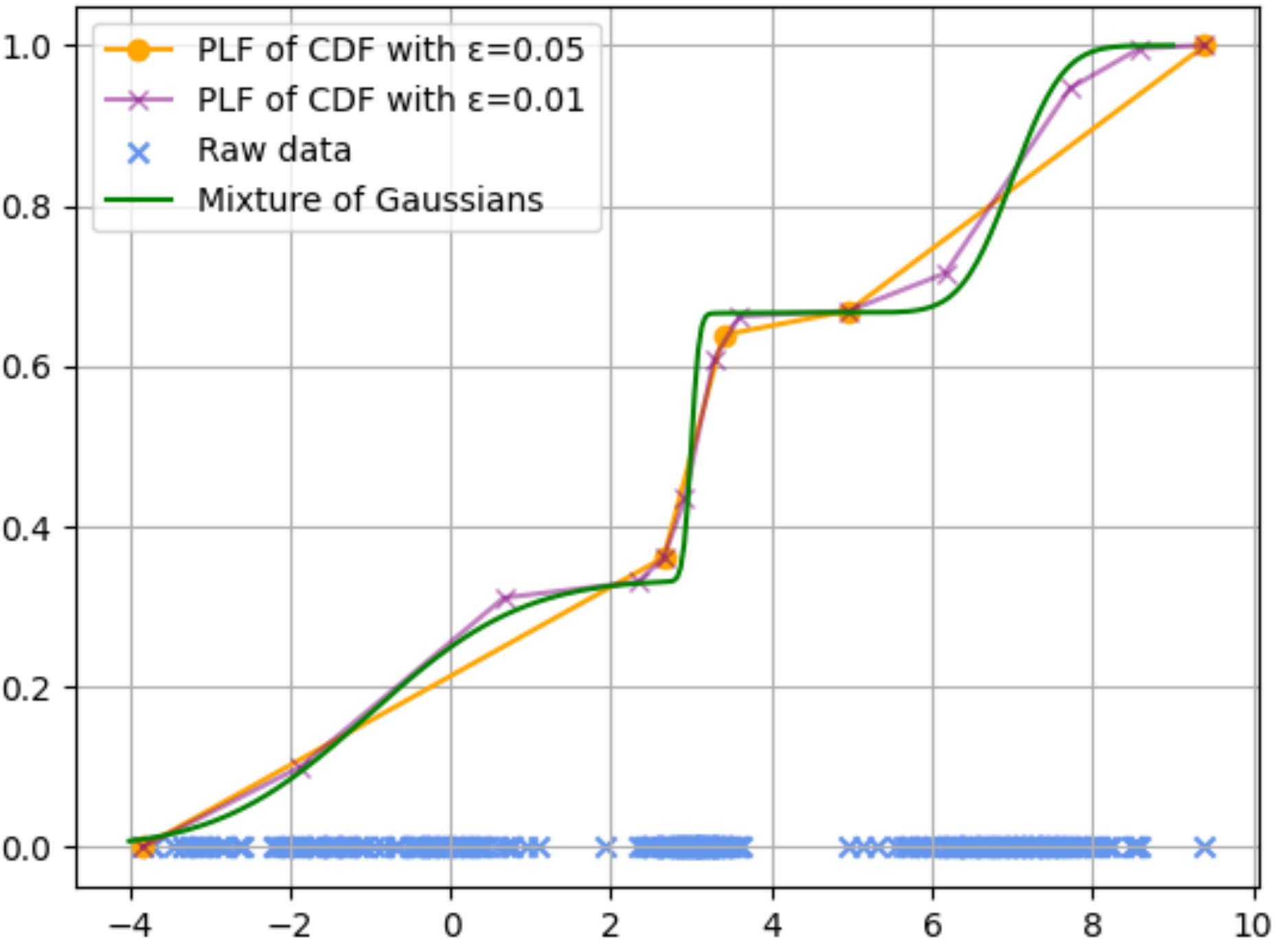} 
    
    \caption{Samples (blue) drawn from a mixture of three Gaussian CDFs 
    (green, log-likelihood of $-640.18$), 
    the CDF approximated by a PLF  with $\epsilon=0.05$ (orange, log-likelihood of $-685.91$) 
    and the CDF approximated by a PLF  with $\epsilon=0.01$. (purple, log-likelihood of $-624.49$) 
    The hinges of the PLF are marked by orange dots and purple crosses.} 
    
    \label{fig:quantile-dist-example}
    
\end{figure}

    
    
    

For several reasons the use of a PLF as a regressor of the CDF of a 
probability distribution is appealing. (1) the concatenation of linear 
functions is very general, so that a PLF is capable of approximating 
functions of arbitrary shape and precision, and thus the resulting CDF 
can be considered free of model assumptions. (2) computations, 
manipulations and interpretations are intuitive and simple, and (3), 
there are efficient algorithms to acquire PLFs, such as 
MARS~\citep{friedman91mars}. And (4) individual pieces can be learnt 
independently on partitions of the data and composed afterwards to form 
the final distribution. Figure~\ref{fig:quantile-dist-example} shows an 
example of a mixture of three Gaussian distributions and the 
approximation of its CDF in the form of a PLF that has been acquired 
from samples drawn. The Figure shows that the PLF is not only similar 
to the CDF of the three Gaussian distributions, but also competitive in 
terms of the likelihood. 

\subsection{Efficient Learning of Cumulative Distributions}
\label{sec:subsymlearning}

In this section, we introduce an efficient algorithm for fitting CDFs 
in the form of PLFs. It is based on recursive partitioning of the data 
set $\Dtilde$ and inspired by regression tree 
learning~\citep{breiman2017cart}. Starting with 
$\Dtilde$, a point $\left<d_{i^*},\gamma_{i^*}\right>\in\Dtilde$ is 
determined, which minimizes the mean squared error (MSE) of a PLF of 
the form

\vspace{-1ex}\begin{align}
    f(d)=\left\{\begin{array}{ll}f_1(d) & \text{ if } d_1 \leq d \leq d_{i^*}\\f_2(d) & \text{ if } d_{i^*} \leq d \leq d_N\end{array}\right.,
\end{align} for all $d\in\Dtilde$. This process is recursively repeated 
on the subsets $\Dtilde_1=\left\{d_1,\ldots,d_{i^*}\right\}$ and 
$\Dtilde_2=\left\{d_{i^*},\ldots,d_N\right\}$ until an error bound 
$\varepsilon$ is being undercut. Note that, if $\varepsilon=0$, the 
process will terminate when all points in $\Dtilde$ have been selected 
as an optimal split point once. In this case, the learnt PLF reaches 
the highest possible likelihood, where every point is perfectly 
matched. The set of all points represents the interval boundaries of 
the PLF. The \forward algorithm is listed in 
Algorithm~\ref{algo:forward}. MSE$(\Dcal)$ determines a function that 
returns the MSE of all points in $\Dcal$ applied to a linear function 
through the minimal and maximal points in $\Dcal$. The \forward pass 
yields the hinge points in $\Dtilde$ that can be connected to form a 
linear spline of the CDF. The CDF is constantly 0 for all $d<\min\Dcal$ 
and constantly 1 for all $d\geq\max\Dcal$.


\begin{algorithm}[tb]
  \small
  
  \caption{\forward}
  \label{algo:forward}

  \begin{algorithmic}[1]
    \Input $\Dtilde=\left\{\langle d_i,\gamma_i\rangle\right\}$, a sorted set of $\gamma_i$-quantiles
    \Output a set of hinge points of a PLF
    \If{MSE$(\Dtilde)<\varepsilon$}
        \State\Return{$\emptyset$}
    \EndIf
    \State Determine \vspace{-2ex}$\Dtilde^{i^*}=\left\langle\Dtilde_1^{i^*},\Dtilde_2^{i^*}\right\rangle$, with $$\Dtilde_1^{i^*}=\left\{d_1,\ldots,d_{i^*}\right\},\Dtilde_2^{i^*}=\left\{d_{i^*},\ldots,d_N\right\},$$ \\\ \ \ such that ${i^*}=\argmin_{i}{\E{\text{MSE}\left(\Dtilde^{i^*}\right)}}$
    \State\Return{$\{i^*\}\cup\forward(\Dtilde_1^{i^*})\cup\forward(\Dtilde_2^{i^*})$}
  \end{algorithmic}
\end{algorithm}



\subsection{Reasoning about Cumulative Distributions}
\label{sec:reasoning}

\textbf{A-priori reasoning} In the previous section we introduced QPDs 
as a model-free representation of probability distributions and 
outlined the advantages of learning the respective CDF. Representing 
the CDF $F$ directly is advantageous over using the PDF, as marginal 
probabilities $\P{X\leq x_0}$ and $\P{X > x_0}$ can be obtained 
directly by evaluating $F$ without the computationally expensive step 
of integrating the PDF. Prior probabilities over intervals $[x_l, x_u]$ 
can be computed by $\P{x_l \leq X \leq x_u} = F(x_u) - F(x_l)$. 

\textbf{A-posteriori reasoning} The calculation of a posterior 
$\Pcond{X}{x_l \leq X \leq x_u}$ can be implemented in a similarly 
straightforward fashion. By decomposing the piecewise linear CDFs 
according to the posteriors' condition, the distribution now represents 
only the required interval $[x_l, x_u]$.
Simply cropping and extracting a part of 
the function at the interval boundaries
would leave us with an invalid QPD, since in most cases it will not 
represent probabilities ranging from $0.0$ to $1.0$. To normalize the 
distribution, the cropped part of the function is shifted to the base 
axis and stretched such that the properties of a valid 
distribution function are restored, i.e. $F(x_l)$ and $F(x_u)$ will 
evaluate to $0.0$ and $1.0$, respectively.

    

\textbf{Confidence-rated output} In many practical applications, 
obtaining a mere predictive value of a target variable is insufficient. 
Especially in safety-critical applications, it is crucial that 
predictions can have a confidence value attached expressing their 
dependability. For example, it is useful to report a confidence 
interval that encompasses the expectation of a variable $X$, 
$\mathds{E}(X)$, with a certain probability -- the confidence level. In 
order to compute such a confidence interval $[x_l,x_u]$ given a confidence 
level $\vartheta$, the inverse of the CDF, also called \textit{percent 
point function} (PPF) can be used,

\begin{align}
        F^{-1}\left(F\brackets{\mathds{E}(X)}-\frac{\vartheta}{2}\right)\leq\mathds{E}(X)\leq F^{-1}\left(F\brackets{\mathds{E}(X)}+\frac{\vartheta}{2}\right).
\end{align} In the case of PLFs, inverting the CDF is cheap and 
involves only the inversion of every linear component of the CDF. In 
the \nameref{sec:exp} section, we present an example of 
confidence-rated outputs in JPTs.

\subsection{Learning and Reasoning in Symbolic Domains}
\label{sec:symreason}

Probability distributions over symbolic variables are represented by 
histograms over the domains of the respective variables. If a path from 
the root of the tree to a leaf node contains a decision node 
constraining a symbolic variable, it is superfluous to store prior 
distributions over the respective variables in the leaves of the 
respective subtree, the impurity (entropy) of the variable is minimal 
already. In order to compute the leaf-conditionals 
$\Pcond{q_i}{\lambda,e_i}$ in Eq.~\ref{eq:jpt-posterior} for all 
symbolic variables $X_i$, it is sufficient to eliminate the 
inadmissible values of $X_i$ from the respective domain and 
re-normalize the histogram distribution.

%% file: experiments.tex
\begin{figure}[t]
    \centering
    \includegraphics[width=\columnwidth]{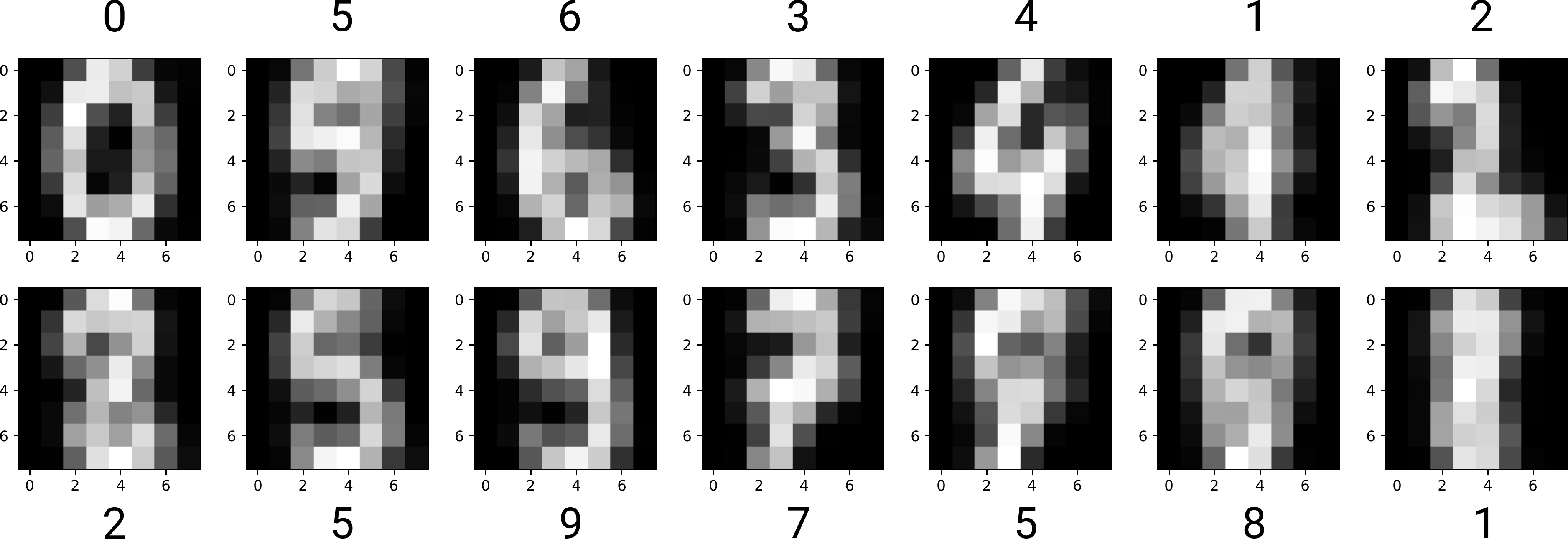}

    \caption{Expectations $\mathds{E}(X_1),\ldots,\mathds{E}(X_{64})$ 
    over the probability distributions in the 14 leaf nodes of a learnt 
    JPT (minimum 100 samples in each leaf). The clusters 
    represented by the images in the leaf nodes reasonably match the 
    associated class labels, which are displayed above the upper row 
    and below the bottom row.}

    \label{fig:mnist}
\end{figure}

In recent years, various approaches for representing and learning 
probability distributions have been proposed, each of which is based on 
a specific set of constraints and assumptions (cf. Section Related 
Work). It is therefore difficult if not impossible to provide 
reasonable and robust evaluation criteria for comparing the predictive 
performance of a joint distribution learnt with different models. 

Likelihood, on the one hand, is a well-known measure for a model's 
ability to fit a data set. However, computing the exact likelihood in 
most previous works strongly depends on the model assumptions made and 
the preprocessing of the data sets. As -- to the best of our knowledge 
-- \jpts are the first approach that allows for hybrid 
symbolic/continuous model-free, joint probability distributions without 
prior assumptions, a direct comparison with previous 
works~\citep{gens2013learning,dang2020strudel,vergari2021compositional} 
is difficult if not impossible.


We additionally evaluate \jpts by comparing their predictive performance 
with respect to each variable in an experiment to a discriminative 
approach that has been trained on the respective variable exclusively 
with the same representational complexity (e.g. the number of samples 
in a leaf node). This is a relatively hard setup, because the 
\textit{single} JPT model is to compete with specialized discriminative 
models, which are allowed to tailor their representational resources to 
one particular variable.




\paragraph{MNIST Data Set}


Another popular dataset for ML model evaluation is known as MNIST 
\citep{lecun-mnisthandwrittendigit-2010}. This dataset contains images 
of the digits 0 to 9, as written by various people. The images' 
dimensions are 8$\times$8 and have grayscale values in the range 0-255. 
Traditionally, the task in this dataset is to correctly assign one of 
the labels 0-9 to every image in the collection, which is a 
discriminative problem statement. In this experiment, we demonstrate 
that JPTs can perform both discriminative classification tasks and 
generative sample generation. The learnt JPT comprises 14 leaves, whose 
expectation over the 65 variables are shown in Figure~\ref{fig:mnist}. 
It can be seen that the expected value over the pixel values reasonably 
matches the expectation over the class labels. A visualization of the 
tree structure itself can be found in the appendix.


\paragraph{Regression}
\label{sec:regression}
\begin{figure}[t]
    \centering
    \includegraphics[width=\columnwidth]{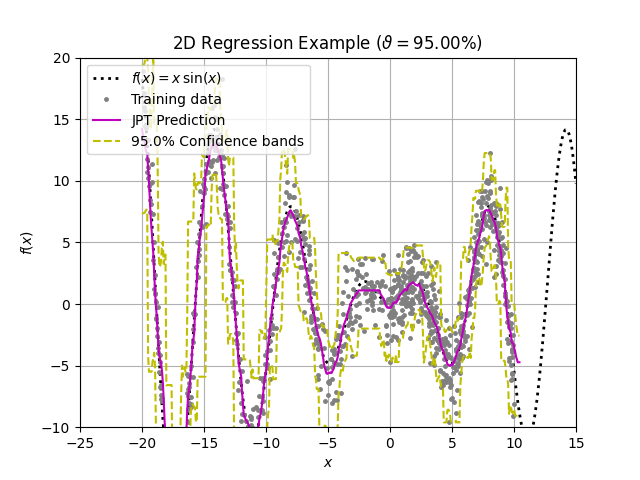}

    \caption{Regression fitting the function $f(x)=x\sin 
    x$: The ground truth function $f(x)$ is shown in dotted 
    gray, the training samples with Gaussian noise as gray dots, 
        the prediction $\mathds{E}(y\! \mid\! x,\vartheta\!=\!.95)$ given by the 
    learnt JPT in purple, and the yellow dashed lines represent the 
    upper and lower percentiles of the prediction.}

    \label{fig:regression-plot}
\end{figure}

In this experiment, we demonstrate that JPTs can be used to accurately 
perform non-linear regression analyses, for which traditionally popular 
methods like ordinary least squares~(OLS), regression trees~(RT), or 
neural network models are chosen. We chose as ground truth regressor 
function $f(x)=x\sin x$, from which we sampled 1000 data points 
uniformly distributed with additive Gaussian noise. Regression analysis 
with joint distributions in JPTs can be achieved by evaluating the 
expected value of $y$ given the value of $x$, $\mathds{E}(y\!\mid\! 
x,\vartheta)$, as an estimate of $f(x)$, where $\vartheta$ is a 
confidence level that is used to calculate the confidence bounds of the 
answer as described in Section~\ref{sec:reasoning}. 
Figure~\ref{fig:regression-plot} shows the ground thruth, the sampled 
training data, the JPT prediciton as well as its upper and lower 
confidence bands. A quantitative comparison to CART is shown in 
Table~\ref{tab:regression-results}.

\begin{table}[t]
    \centering
    \resizebox{6cm}{!}{
    \begin{tabular}{p{4cm}rrrr}
    \hline
      \textbf{\# samples.}   &   \textbf{JPT}    &   \textbf{CART} \\
    \hline
      20.0\%   & 3.0041032  & 4.821578 \\
      10.0\%  & 2.1888970 & 4.0494846 \\
      5.00\%  & 1.6213786 & 2.3802814 \\
      2.00\%  & 1.2783260 & 1.3080680 \\
      1.00\%  & 0.7854674 & 0.9564107 \\
    \hline 
    \end{tabular}}
    
    \caption{MAE (lower is better) for predictions in 
    the regression experiment using JPT and CART with different 
    learning parameters. `\# samples' denotes the fraction of available 
    data samples that need to be covered by every leaf node.}

    \label{tab:regression-results}
\end{table}


\paragraph{Airline Departure Delay}\label{exp:airline}
\jpts can efficiently be learnt even with very large datasets. As an 
example, we are using the publicly available \textit{Airlines Departure 
Delay Prediction} dataset provided by OpenML~\citep{OpenML2013}. This 
dataset is particularly of interest, as it comprises 10 million 
instances each of which consists of 7 mixed-type feature values (3 
nominal, 4 numeric). There are no missing values in that dataset. The 
results in Table~\ref{tab:airline-results} show that even though the JPT 
model represents all variables as opposed to the specialized CART 
models each of which only represents one variable, the JPT performs 
comparably well on all of the 7 variables and even outperforms the CART 
models in three of them. This is remarkable, because it shows, that JPTs
can compete with specialized discriminative models.  

\begin{table}[b]
    \centering
\resizebox{.9\columnwidth}{!}{
\begin{tabular}{p{3cm}rrrr}
\hline
\textbf{Num. Variable}   &   \textbf{JPT} &     (+/-) &   \textbf{CART} &      (+/-) \\
\hline
DayOfWeek  &   1.89726 & 0.0118047 &   1.70767 & 0.00376937 \\
CRSDepTime & 144.518   & 0.615262  & 163.463   & 0.550986   \\
Distance   & 293.35    & 2.82525   & 397.128   & 2.83425    \\
CRSArrTime & 139.678   & 0.825618  & 172.927   & 0.6609     \\
\hline
\end{tabular}}
\\
\resizebox{.9\columnwidth}{!}{
\begin{tabular}{lrrrr}
\hline
\textbf{Sym. Variable}      &    \textbf{JPT} &       (+/-) &    \textbf{CART} &       (+/-) \\
\hline
UniqueCarrier & 0.178936  & 0.000313542 & 0.200474  & 0.000179531 \\
Origin        & 0.0639007 & 3.28761e-05 & 0.0967244 & 1.47472e-05 \\
Dest          & 0.0629808 & 1.9464e-05  & 0.0966721 & 1.41371e-05 \\
\hline
\end{tabular}}

\caption{Experimental results on the Airline dataset: MAE (lower is 
better) and deviations of the numeric variables of JPT and CART (top) 
and F-score (higher is better) and deviation of the symbolic variables 
(bottom).}

\label{tab:airline-results}
\end{table}

\paragraph{Empirical Evaluation}

We conducted an extensive evaluation of JPTs on eight popular datasets 
from the UCI machine learning repository~\citep{Dua:2019}, in which we 
learnt JPTs with different hyperparameters and determined the size of 
the resulting models as well as the likelihoods they achieve. As a 
hyperparameter for learning, we varied the minimum number of samples in 
each leaf node to different portions of the available training data to 
induce different levels of model complexity. The evaluation has been 
conducted on test sets created by randomly sampling 10\% of the data. 
Our experiments show that, with increasing complexity (\ie decreasing 
minimum number of samples per leaf), the achieved likelihood of the 
learnt models reliably increases significantly both in the training set 
and in the test set. The detailed experimental results are listed 
in Table~\ref{tab:empirical} in the appendix. The rightmost column
contains the number of test samples with 0 likelihood. This typically
happens in model-free distributions where test samples may lie outside the
convex hull of the training data and the \textsc{CDF-Learn} algorithm
assigns 0 probability mass to these regions.

%% file: desiderata.tex
Joint probability distributions have great potential to serve as 
powerful problem-solving tools for machine learning applications. As 
opposed to most methods in the field of ML, joint distributions do not 
rely on dedicated input and output variables but can be queried for any 
aspect contained in the model given any evidence. Most existing 
 methods, however, require strong assumptions that all 
variables are either symbolic or numeric or that their distributions 
have the same functional form. This makes the accompanying methods 
tailored to the specific densities. However, in many real-world 
problems, the semantics of variables calls for support of heterogeneous 
modalities of both numeric and categorical variables in a single model. 
JPTs meet this requirement of hybrid symbolic and subsymbolic 
reasoning, which has also been identified as one of the key demands of 
future AI applications~\citep{marcus2019rebooting}. A further 
significant advantage of JPTs over state-of-the-art PGMs is that JPTs 
do not rely on assumptions about dependencies or families of 
distributions at all. The variable dependencies are represented by the 
model structure that is generated as a byproduct of the learning 
procedure and the use of piecewise linear functions to approximate the 
CDFs of numeric variables allows highest flexibility with minimal model 
assumptions. The tree structure itself consists of conjunctions of 
variable constraints, each of which describes a specific subregion in 
the problem space. This makes the model structure interpretable and 
understandable for humans, and, as a consequence, its allows 
transparent reasoning and more robust, more traceable, and more 
explainable decision making~\citep{goebel2018explainableai}. The 
practical applicability of classic PGM is impeded by their 
representational and computational complexity that the models typically 
imply. The inferential complexity is $\#P$-complete in the general 
case~\citep{koller2009probabilistic} and learning is intractable since 
it involves inference. To circumvent these computational challenges, 
strong assumptions about variable independence and approximate 
inference mechanisms have to be adduced. In JPTs, inference is exact 
and can be performed in linear time with respect to the model size, 
i.e. the number of leaves in the tree. In addition, it is easy to limit 
the complexity of the model by, for instance, the computational 
resources that are available. This makes JPTs extremely flexible, 
scalable and parallelizable. To the best of our knowledge, \jpts
are the only shallow \textit{deterministic} probablistic circuit, such 
that the circuit remains tractable for MPE inference.

In summary, we argue that JPTs address a selection of properties of 
machine learning methods that have been identified as pivotal 
challenges of AI and ML in the literature of recent years. In this 
section, we reviewed a selection thereof and discussed in a qualitative 
manner, in which way JPTs are capable of addressing these desiderata. 
For a more detailed discussion of activities and methods in a typical 
lifecycle of ML models we refer to the survey by 
\cite{ashmore2021assuring}.

%% file: related-work.tex

The field of probabilistic learning and reasoning is an important and 
actively investigated field of AI, and many works have been presented 
in the past years, which aim at making probabilistic models more 
tractable. \cite{poon2011sum} represent the partition function of PGMs 
by introducing multiple layers of hidden variables by a deep 
architecture called sum-product networks (SPN). SPNs are rooted 
directed acyclic graphs, whose root values represent an unnormalized 
probability distribution and can be learnt recursively by either 
splitting it into a product of SPNs over independent sets of variables 
or into a sum of SPNs learned from subsets of the instance, if they 
cannot be found. SPNs perform approximate inference in the case of 
cyclic dependencies. Otherwise cycles have to be replaced by 
multivariate distributions whose representation is generally 
untractable \citep{gens2013learning}. As opposed to SPNs, JPTs do not 
rely on a dependency model prior to learning and support explainable 
reasoning in both symbolic and continuous variable domains. Contrasted 
with probabilistic circuits (PC), JPTs' dependency model is learnt 
automatically from data and can express distributions without any model 
assumptions~\citep{shah2021piu}. In general, all PGMs postulate a rigid 
dependecy model that must be known before learning the model parameters 
and learning the structure of such a model imposes an additional 
NP-hard problem. In JPTs, the dependency model is automatically learned 
from the data while no model needs to be specified in advance. Like 
\jpts, cutset networks~\citep{rahman2014cutset} do recursive 
partitioning, but fit arbitrary Bayesian networks in the partitions. 
Cutset networks, however, are bound by the complexity of the Bayesian 
networks in the leaves and lack the ability to generally represent 
continuous variables. JPTs thus can be considered generalizations of 
both SPNs, PCs and CNs without the necessity to pre-define the model 
structure. \cite{bishop1994mixture} introduces Mixture Density Networks 
(MDNs), which combine conventional neural networks with mixture density 
models. MDNs provide a general framework modeling conditional 
probability densities of the output variables, thus enabling a 
multi-valued mapping. Similarly to JPTs, MDNs are mixture models 
representing distributions over symbolic and subsymbolic variables. 
However, in contrast to MDNs, JPTs are generative models representing 
joint distributions over all variables instead of conditional 
probability densities of the output variables, which limits MDNs to be 
discriminative. Additionally, JPTs profit from the benefits of 
white-box tree structures that are intuitively interpretable, offer 
explainable decisions and allow efficient learning and reasoning. 
\cite{bishop2012bayesian} use variational inference on Hierarchical 
Mixture of Experts (HME) to provide a Bayesian treatment of the model. 
It can be used to solve multi-dimensional regression problems and is 
expected to also work on binary classification problems. The HMEs often 
find local maxima which are not necessarily good solutions which the 
authors try to overcome by using multiple runs to select the best 
optima. However, this may be infeasible for large datasets. For a more 
detailed review of mixture-of-expert models we refer to 
in~\citet{yuksel2012twenty}.

%% file: conclusions.tex
We introduced Joint Probability Trees~(JPT), which are tree-based 
representations that allow the compact representation and efficient 
learning of and reasoning about joint probability distributions. As 
opposed to PGMs, JPTs support learning and reasoning in hybrid domains, 
i.e. they allow to coalesce numeric and symbolic variables in one 
single model in a sound and consistent way. A key feature of JPTs is 
that the dependencies among variables are represented in a tree 
structure, whose leaves maintain univariate prior distributions over 
all variables. For numeric variables, we propose quantile-parameterized 
distributions, which can be regarded as model-free representations of 
arbitrary numeric distributions. Our experiments show that JPTs are 
able to accurately learn and represent complex interactions between 
many variables, while keeping interpretability and scalability. We 
argue that the challenges that we address with JPTs will be key in 
making the transition from narrow, specialized expert systems to 
hybrid, high performance AI systems and in pushing todays computer 
systems towards more generally intelligent and more robust and reliable 
decision making.

%% file: appendix.tex


\subsection{Empirical Evaluation}

\begin{table}[h]
\centering
\resizebox{.935\textwidth}{!}{
 \begin{tabular}{p{5cm}|c|c|c|c|c}
     \toprule
      \textbf{Dataset} & \textbf{Min samples per leaf} & \textbf{Model Size} &  \textbf{Average Train Log-Likelihood} & \textbf{Average Test Log-Likelihood} &  \textbf{0-likelihood Test Samples} \\
     \hline\hline
                             
    \multirow{3}{*}{\shortstack[l]{\textbf{IRIS Dataset}\\Examples:  150\\Variables:  5}} 
             &                        90\% &                   23 &                 -5.45 &   -5.63 &                        - \\
             &                        40\% &                   46 &                 -3.54 &  -3.33 &                        - \\
             &                        20\% &                   94 &                 -2.16 & -2.66 &                        20\% \\
             &                        10\% &                  184 &                 -1.37 & -1.91 &                        27\% \\
             &                       5\% &                  353 &                 -0.18 & -1.2 &                        67\% \\
             &                       1\% &                  679 &                  6.11 &                 - &                        100\% \\\hline
    \multirow{3}{*}{\shortstack[l]{\textbf{Adult Dataset}\\Examples:  32561\\Variables:  15}}
     &                        90\% &              1472557 &                -55.46 &  -54.98 &                        49\% \\
     &                        40\% &              2945118 &                -53.79 &   -53.07 &                        58\% \\
     &                        20\% &              5890233 &                -52.96 & -52.45 &                        64\% \\
     &                        10\% &             11780477 &                -51.17 &  -50.18 &                        70\% \\
     &                       5\% &             22088384 &                -50.13 & -50.25 &                        81\% \\
     &                       1\% &             94243769 &                -41.57 & -42.96 &                        93\% \\\hline
    \multirow{3}{*}{\shortstack[l]{\textbf{Dry Bean Dataset}\\Examples:  13611\\Variables:  17}}
     &                        90\% &               476858 &                -17.49 & -16.98 &                        95\% \\
     &                        40\% &               953709 &                -12.13 & -13.23 &                        95\% \\
     &                        20\% &              1430565 &                 -9.80 & -10.68 &                        95\% \\
     &                        10\% &              3814841 &                 -4.91 &  -6.75 &                        95\% \\
     &                       5\% &              7629662 &                 -2.31 & -4.11 &                        95\% \\
     &                       1\% &             36717612 &                  2.41 &  0.85 &                        96\% \\\hline
     \multirow{3}{*}{\shortstack[l]{\textbf{Wine Dataset}\\Examples:  178\\Variables:  13}}
     &                        90\% &                 1550 &                -18.97 & -19.83 &                        56\% \\
     &                        40\% &                 3101 &                -17.06 &  -18.69 &                        83\% \\
     &                        20\% &                 6197 &                -13.42 & -14.9 &                        94\% \\
     &                        10\% &                12417 &                -10.98 & -12.46 &                        94\% \\
     &                       5\% &                24824 &                 -6.74 &                 - &                        100\% \\
     &                       1\% &               207028 &                 41.22 &                 - &                        100\% \\\hline
     \multirow{3}{*}{\shortstack[l]{\textbf{Wine Quality Dataset}\\Examples:  6497\\Variables:  13}} 
     &                        90\% &                   67 &                 -9.50 &  -9.8 &                        - \\
     &                        40\% &                  138 &                 -8.10 &  -8.34 &                        1\% \\
     &                        20\% &                  204 &                 -7.60 &  -7.68 &                        1\% \\
     &                        10\% &                  479 &                 -6.48 &  -6.57 &                        2\% \\
     &                       5\% &                 1072 &                 -5.67 &  -5.85 &                        4\% \\
     &                       1\% &                 5307 &                 -3.08 & -3.82 &                        12\% \\\hline
     \multirow{3}{*}{\shortstack[l]{\textbf{Bank and Marketing Dataset}\\Examples:  45211\\Variables:  17}}
     &                        90\% &               116433 &                -34.60 & -34.64 &                        07\% \\
     &                        40\% &               232866 &                -34.50 &    -34.54 &                        8\% \\
     &                        20\% &               465732 &                -34.31 &  -34.32 &                        12\% \\
     &                        10\% &               815031 &                -33.00 &  -32.77 &                        20\% \\
     &                       5\% &              1630062 &                -32.43 &  -32.08 &                        32\% \\
     &                       1\% &              7335279 &                -30.46 &  -29.45 &                        73\% \\\hline
      \multirow{3}{*}{\shortstack[l]{\textbf{Car evaluation Dataset}\\Examples:  1728\\Variables:  6}} 
     &                        90\% &                   21 &                 -6.89 &  -7.03 &                        - \\
     &                        40\% &                   21 &                 -6.89 &  -7.03 &                        - \\
     &                        20\% &                   63 &                 -6.52 &   -6.62 &                        - \\
     &                        10\% &                  147 &                 -6.48 &  -6.61 &                        - \\
     &                       5\% &                  231 &                 -6.45 &  -6.59 &                        - \\
     &                       1\% &                 1249 &                 -6.38 &  -6.66 &                        - \\\hline
      \multirow{3}{*}{\shortstack[l]{\textbf{Abalone Dataset}\\Examples:  4177\\Variables:  9}}
         &                        90\% &                   64 &                  0.09 & -0.04 &                        - \\
         &                        40\% &                  134 &                  3.65 &  3.66 &                        - \\
         &                        20\% &                  208 &                  5.34 &   5.11 &                        - \\
         &                        10\% &                  482 &                  8.14 &      8.05 &                    2\% \\
         &                       5\% &                 1034 &                  9.32 &   9.28 &                        3\% \\
         &                       1\% &                 5502 &                 11.34 &  10.74 &                        18\% \\\hline
\end{tabular}}

\caption{Results of the evaluation of JPTs on eight benchmark data sets 
of the UCI machine learning repository~\citep{Dua:2019} for different 
hyperparameters. ``Min samples per leaf'' means that at least the 
respective percentage of data points available for training must be 
represented by any leaf of the tree. ``Min samples per leaf''=90\% thus 
results in a JPT with only one leaf, \ie a set of independent prior 
distributions over all variables considered. ``0-likelihood test 
samples'' determines the percentage of test samples with 0 likelihood. 
This may happen to samples lying outside the convex hull of the 
training data, where 0 probability mass is assigned by the 
\textsc{CDF-Learn} algorithm.}

\label{tab:empirical}
\end{table}

\newpage

\subsection{JPT Example - Tree}
\label{app:example}

\vspace{5em}
\begin{figure}[h]
    \centering
    \includegraphics[width=\columnwidth]{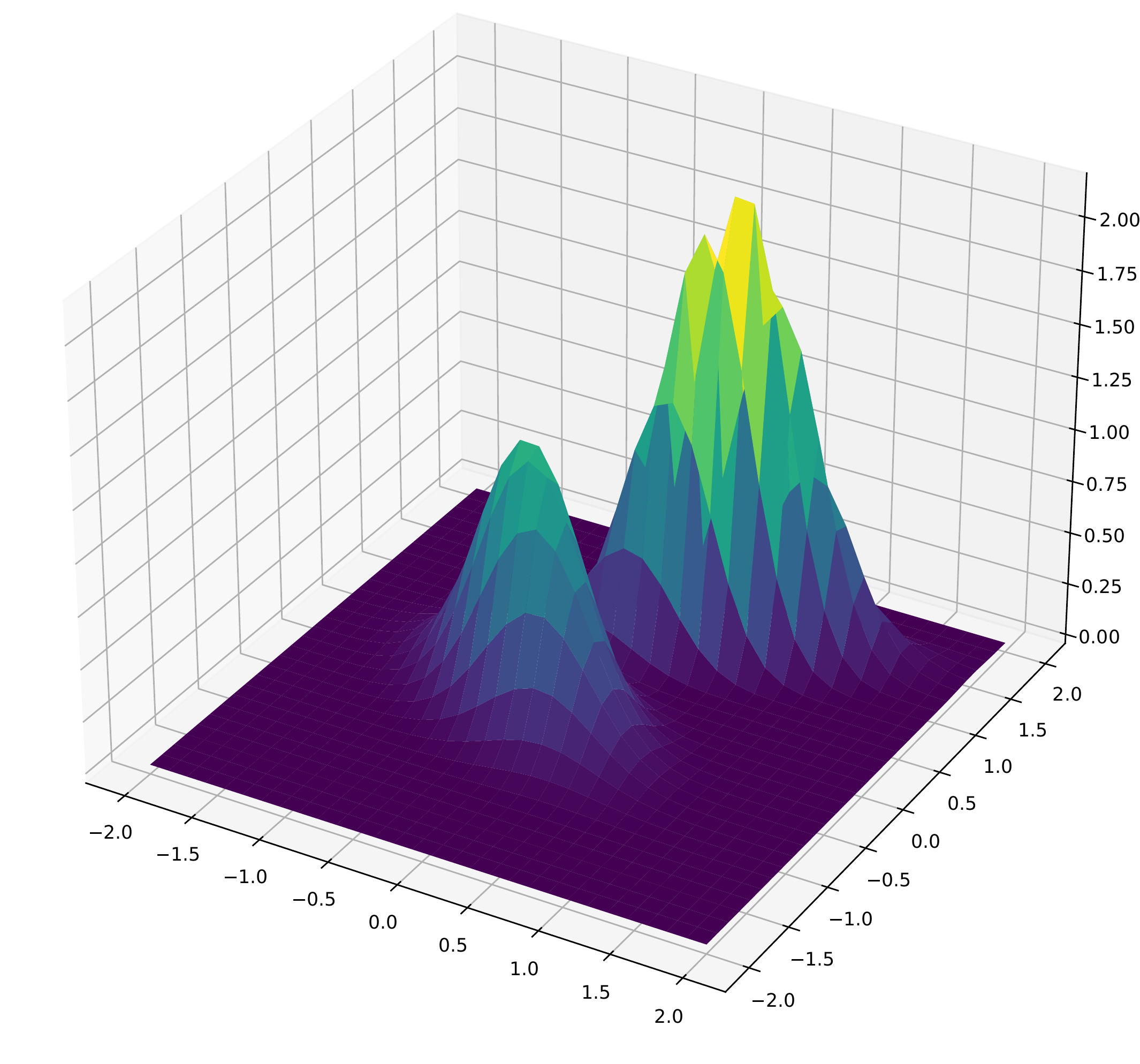}

    \caption{The ground truth distribution of the toy data set in Section~\ref{sec:example}.}

    \label{fig:}
\end{figure}
\newpage

\vspace*{5em}
\begin{figure}[h]
    \centering
    \includegraphics[width=\columnwidth]{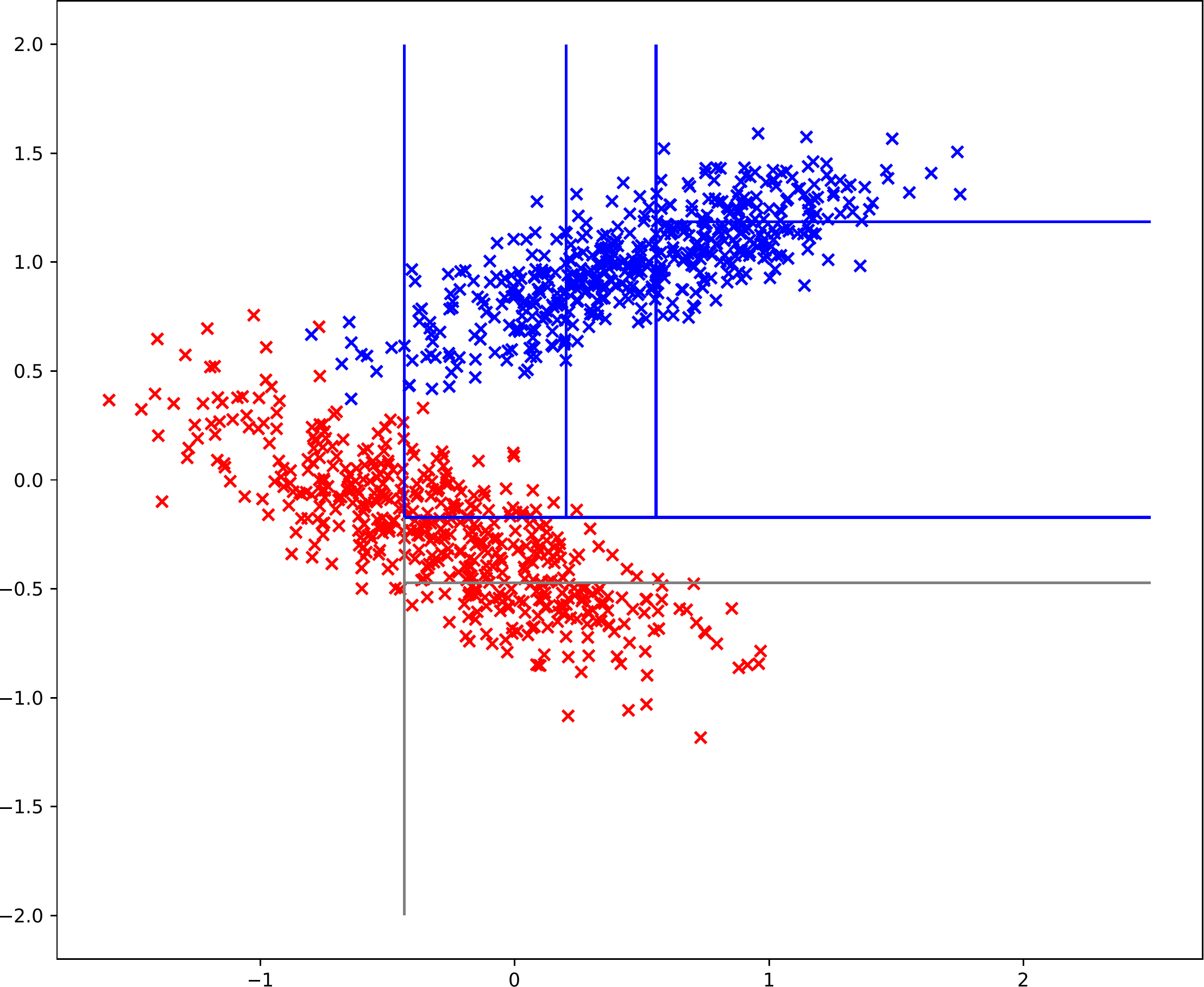}

    \caption{The scatterplot of the underlying toy data set in Section~\ref{sec:example}.}

    \label{fig:}
\end{figure}
\newpage

\vspace*{5em}
\begin{figure}[h]
    \centering
    \includegraphics[width=\columnwidth]{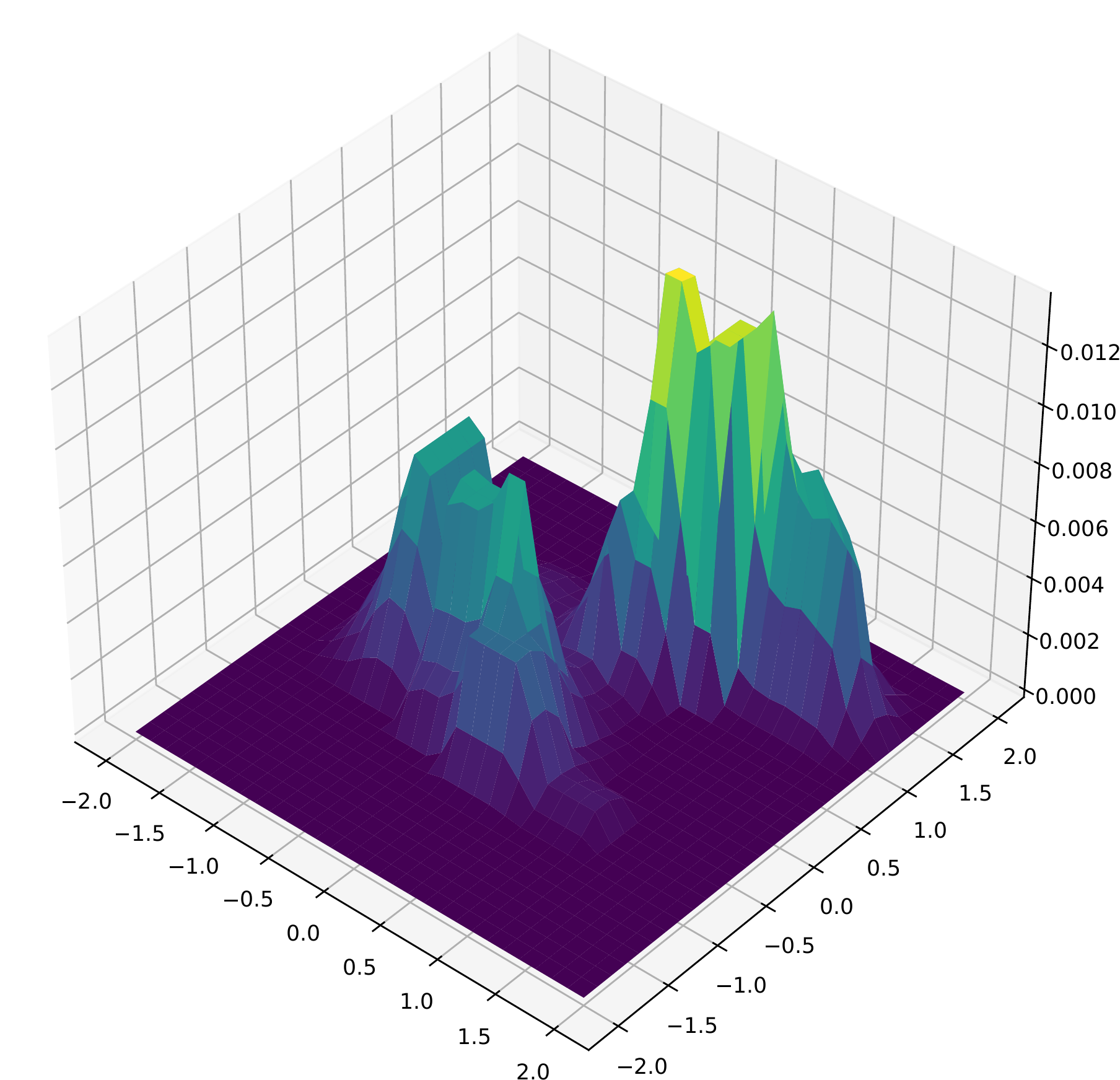}

    \caption{The plot of the marginal joint distribution $P(X,Y)$ of the toy data set in Section~\ref{sec:example}.}

    \label{fig:marginal}
\end{figure}
\newpage

\vspace*{8em}
\begin{figure}[h]
    \centering
    \includegraphics[width=\columnwidth]{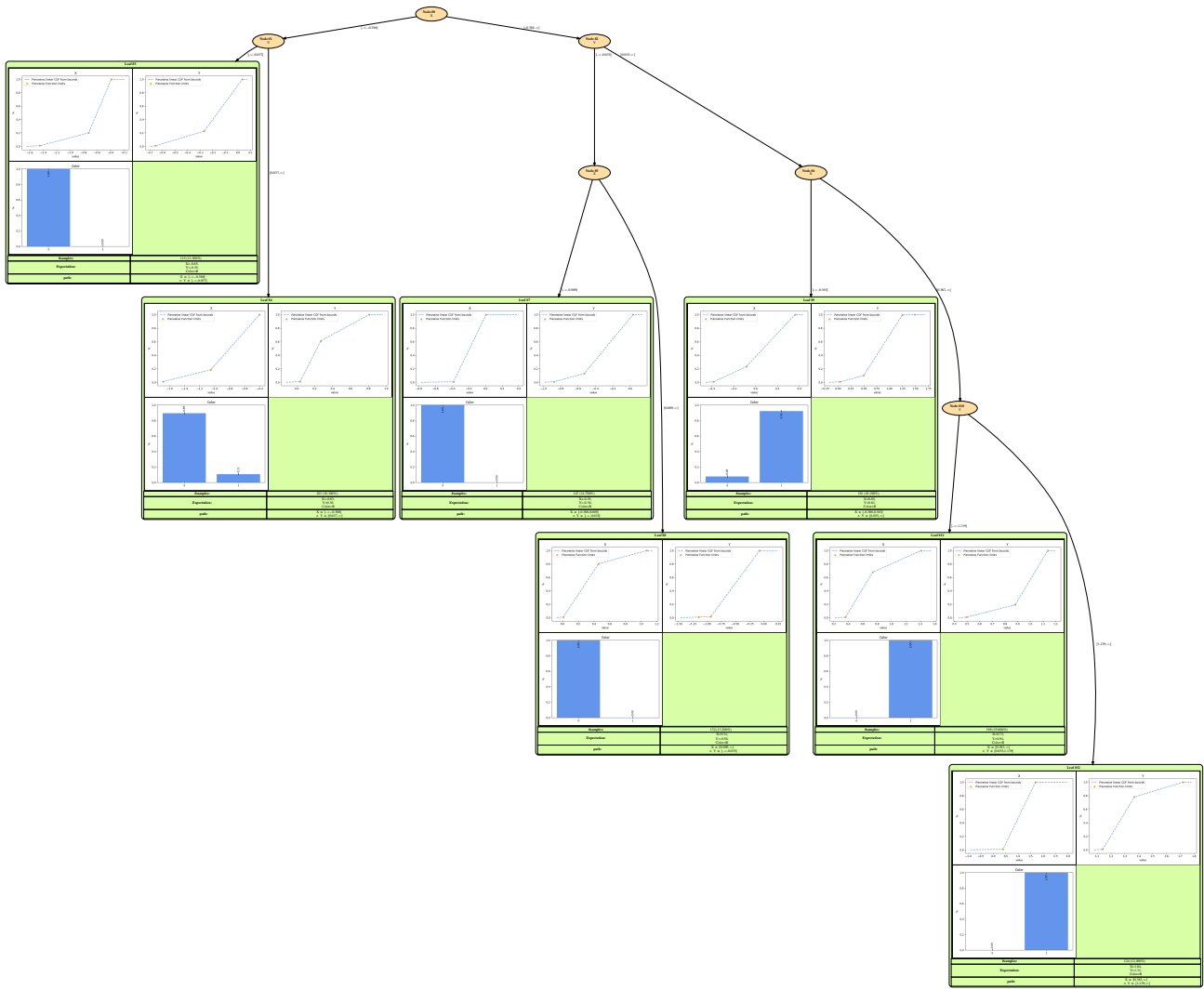}

    \caption{The JPT structure learnt using the toy data set in Section~\ref{sec:example}.}

    \label{fig:guassian}
\end{figure}
\newpage

\subsection{MNIST - Tree}
\label{app:mnist}

\begin{adjustbox}{angle=90,center,caption=The JPT structure learnt using the MNIST data set.,nofloat=figure}
        \includegraphics[height=.8\textwidth]{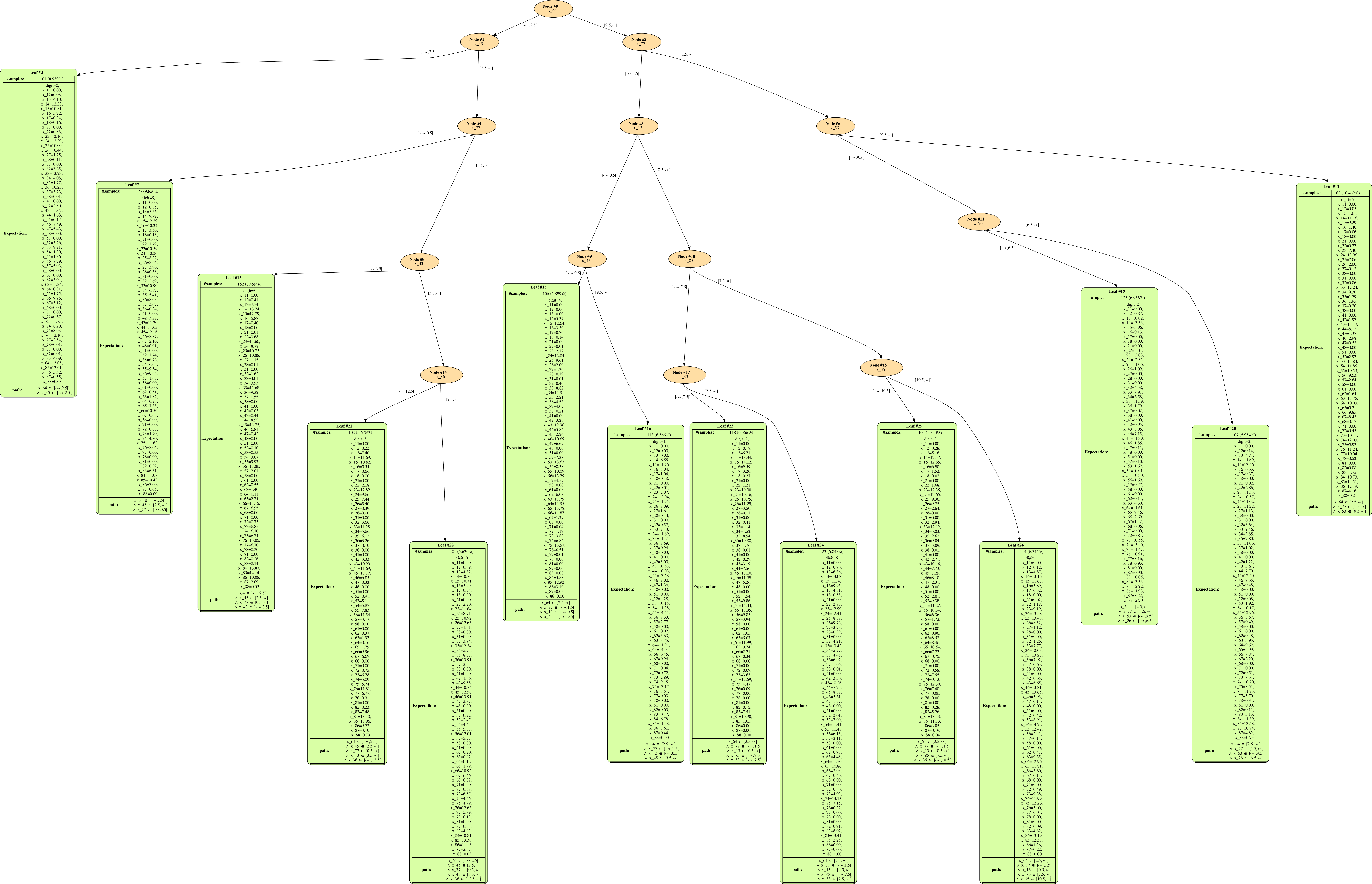}


\end{adjustbox}